\documentclass{article}
\pdfoutput=1
\usepackage[utf8]{inputenc} 
\usepackage[T1]{fontenc}    
\usepackage{hyperref}       
\usepackage{url}            
\usepackage{booktabs}       
\usepackage{amsfonts}       
\usepackage{nicefrac}       
\usepackage{microtype}      

\usepackage{cleveref}
\usepackage{graphicx}
\usepackage{caption}
\usepackage{subcaption}
\usepackage{bm}
\usepackage{color}
\usepackage{natbib}
\usepackage{amsmath,amssymb}
\usepackage{breqn}
\newcommand{\Exv}[2][]{\mathbf{E}_{#1}\left[ #2 \right]}

\usepackage{color}
\usepackage{amsmath}
\usepackage{amssymb}
\usepackage{amsthm}
\usepackage{wrapfig}

\newcommand{\notate}[1]{}

\newtheorem{example}{Example}

\newtheorem{lemma}{Lemma}
\newtheorem{definition}{Definition}

\newif\iflong

\title{Parallel SGD: When does averaging help?}

\author{\hspace{-1.5em}Jian Zhang,
Christopher De Sa,
Ioannis Mitliagkas\thanks{Department of Statistics, Stanford University},
and Christopher R\'{e} \\
Department of Computer Science, Stanford University\\
\texttt{\{zjian,cdesa,imit,chrismre\}@stanford.edu}
}

\begin{document}

\maketitle

\begin{abstract}
	Consider a number of workers running SGD independently on the same pool of data and averaging the models every once in a while --- a common but not well understood practice. We study model averaging as a variance-reducing mechanism and describe two ways in which the frequency of averaging affects convergence.
   For convex objectives, 
   we show the benefit of frequent averaging depends on the gradient variance envelope.
   For non-convex objectives, we illustrate that this benefit depends on the presence of multiple optimal points.
   We complement our findings with multicore experiments on both synthetic and
   real data.

\end{abstract}

\newcommand{\expit}{\bar{w}_k^t}
\newcommand{\expitprev}{\bar{w}_{k-1}^t}

\section{Introduction}
\label{sec:intro}

Stochastic gradient descent (SGD) \cite{shalev2009stochastic,zhang2004solving,zinkevich2003online}
is the workhorse of modern machine learning, largely due to its simplicity and scalability to very large datasets \cite{bottou2010large}.
As a result, there has been a lot of interest in parallelizing SGD \cite{delalleau2007parallel,recht2013parallel,recht2011hogwild,zinkevich2010parallelized}.
One simple method is to run $M$ independent workers in parallel for a number of steps and then average all the models (decision variables) at the end \citep{zinkevich2010parallelized}. This method is often referred to as {\em one-shot averaging}. At the other extreme, one could average the models after {\em every iteration}. When using $M$ workers, this is statistically equivalent to a single worker running SGD with mini-batches of size $M$; we refer to this parallel method as {\em mini-batch averaging}.
One-shot averaging reduces variance only at the end of execution, but requires very little communication. In comparison, mini-batch averaging reduces variance {\em at every step} and therefore converges after fewer steps, i.e.\ it has higher {\em statistical efficiency}. However, it requires much more communication, making each step more expensive computationally, i.e.\ it has lower {\em hardware efficiency}. As a result, we can use the averaging rate to trade off statistical and hardware efficiency in order to maximize the performance, in terms of wall clock time, of the algorithm.

\notate{Skipped statistical efficiency figure here due to space constraints.}

In this paper, we study the method where $M$ independent workers run in parallel and we take an average of their models periodically.
We refer to the periods between averaging operations as {\em phases}. Although this kind of algorithm is used in practice, and some recent results \cite{zhang2014dimmwitted} suggest that more averaging is better, there are no known recipes for averaging that provide theoretical guarantees. How frequently should we average? Does more averaging even necessarily lead to faster convergence? Recently, Bijral et al. \cite{bijral2016data} addressed a similar question in a distributed setting. They show that the rate of averaging affects the convergence of the consensus process on a computation graph.
Our contributions are different in nature:
\begin{itemize}
\item For convex problems, we show that the envelope of gradient variance is important:
when the variance is higher far from the optimum, frequent averaging leads to faster convergence. 
For a simple model, we show that frequent averaging allows convergence to a smaller noise ball.
\item For non-convex problems, we illustrate that one-shot averaging can produce inaccurate results, and show how more frequent averaging can be used to avoid this effect.
\item We implement this averaging scheme on a multicore machine and run a number of experiments. The results support our claims in both the convex and non-convex settings.
\end{itemize}

%
%
%

\section{Formulation}

\newcommand{\reals}{\mathbb{R}}

We aim to minimize the following objective, consisting of $m$ components, with respect to $w \in \mathbb{R}^n$.
\begin{equation}
\min_{w \in \reals^n } f(w) \triangleq  {1 \over m} \sum_{j=1}^m f_j(w)
\label{eqn:optimization}
\end{equation}
Here each $f_j$ is convex, and typically represents an individual example from a dataset. We denote the global optimum by $w^*$.
Stochastic gradient descent draws one of the component functions uniformly at random to perform each iteration. The $k$-th iterate update is described by
\begin{equation}
\label{eqn:sgd}
w_k = w_{k-1} - \alpha\nabla f_{\sigma(k)}(w_{k-1}),
\end{equation}
where $\sigma(k) \sim \text{Uni}(1,\dots,m)$ are independent, uniform draws from $[m]$.

\textbf{Parallelization.} Now consider $M$ worker threads, independently performing the update in Equation~\eqref{eqn:sgd} starting from a common point, $w_{0}$. As described in the introduction, execution is split into a number of phases which are defined by averaging steps: each averaging step marks the end of a phase. Phases last for $K$ steps.
For worker $i \in [M]$ the $k$-th iterate of phase $t$ is given by
\begin{equation}
\label{eqn:workerupdate}
w^t_{ik} = w^{t}_{i(k-1)} - \alpha \nabla f_{\sigma_t(i,k)}(w^{t}_{i(k-1)}),
	\quad k \in [K],
\end{equation}
where $w_{i0}^t=w_0^t$, i.e.\ all workers start the $t$-th phase at the same point.
At the end of each phase we take an average of all models and use it as a starting point of the next phase: $w_{i0}^{t+1} = \bar{w}_K^{t}  \triangleq  {1 \over M } \sum_{i=1}^M w_{iK}^t$.

\subsection{Negative results}
Intuition and practice suggest that more frequent averaging should improve statistical efficiency -- i.e.\  result in convergence in fewer steps. However, this behavior is not empirically observed in all cases. In addition, conventional modeling and analysis may fail to capture the empirical benefits of frequent averaging. Before producing our results, we demonstrate this complication by means of two negative results. 

\textbf{Example 1: Homogeneous Quadratics.}
Consider the case where component functions are quadratics with the same Hessian,
$f_j(w) = {1 \over 2} w^\top P w + w^\top q_j + r_j$.
In this case the gradient is linear: $\nabla f_j(w) = P w + q_j$. This implies that one-shot averaging is {\em equivalent} to mini-batch averaging and any other scheme which interpolates between the two, averaging once in a while. To see this, 
consider mini-batch averaging:
$
	\bar{w}^{MB}_{k} \triangleq \sum_{i=1}^M w_{ik}/M
$ where
$w_{ik} = \bar{w}^{MB}_{k-1} - \alpha \nabla f_{\sigma(i,k)}(\bar{w}^{MB}_{k-1})$.
By linearity of the gradient and homogeneity of the Hessian we get equivalence to one-shot averaging.
This is an example where frequent averaging actually offers no improvement over one-shot averaging.

Next, we show that standard modeling assumptions can also lead to false negative results.
\begin{definition} The variance of a gradient evaluated at $w$ is
\[
\Delta(w) \triangleq {1 \over m} \sum_{j=1}^m \left\|  
					\nabla f_j (w)
				-  \nabla f(w)
\right\|^2.
\]
\end{definition}
\textbf{Example 2: Coarse Modeling.}
Typical analyses of SGD use a uniform gradient variance bound,
$
\Delta(w) \leq \sigma^2, \ \forall w,
$ eg.\ \cite{zinkevich2010parallelized,agarwal2011distributed,dekel2012optimal,bijral2016data}. Using this model and assuming $L$-Lipschitz gradients and $c$-strong convexity, the variance of worker $i$'s model estimate after $k$ steps is bounded by
\begin{equation}
	\label{eqn:coarse-variance}
	\mathbb{E} \| w_{ik} - \bar{w}_k\|^2
	\leq   { \alpha \sigma^2 \over 2 L - \alpha c^2}
	\left[
		1 - \left(
			1 - 2\alpha L + \alpha^2 c^2
		\right)^k
	\right]
		\leq   { \alpha \sigma^2 \over 2 L - \alpha c^2}.
\end{equation}
Averaging only reduces variance. Under coarse modeling, variance is linear with respect to $\sigma^2$ and does not otherwise depend on the sequence of $w_{ik}$. 
This implies that as long as we take an average at the end, earlier averaging steps have no measurable effect: the resulting variance is roughly $\alpha \sigma^2 M^{-1} (2 L - \alpha c^2)^{-1}$ for both one-shot averaging and mini-batch setting. For general homogenous quadratics, we can show the bias term $\|\bar{w}_k - w^{\star}  \|$ is unaffected by averaging frequency. It implies the squared distance to optimum should expect to be roughly the same for all averaging settings. 
Thus the conventional modeling fails to capture the empirical benefits of frequent averaging.

\subsection{Model for Gradient Variance}
\label{sec:variance-model}
We introduce a model for gradient variance, capable of capturing the effects of frequent averaging,
\iflong
\else
\begin{align}
\label{eqn:newmodel}
\Delta(w) \leq&  \beta^2 \left\|  
				w -  w^* \right\|^2
				+ \sigma^2.
\end{align}
\fi
The new $\beta$-dependent term causes early-stage benefits for averaging, because new variance introduced at every step now depends on the distance from the optimum, $w^*$. Taking an averaging step early on will reduce this distance and thereby reduce variance throughout execution, leading to convergence in fewer steps.
This phenomenon is stronger the farther we are from the optimum; we expect that speedups 
will mostly be observed when the first variance term dominates, i.e., $\rho \triangleq \beta^2 \|w_0 - w^*\|^2/\sigma^2$ is large. In Section~\ref{sec:experiments}, we will show experimentally that the averaging speedup correlates with the magnitude of $\rho$.

\subsection{Stochastic Averaging Analysis}
\label{sec:stochastic-avg}

We study how the rate of averaging affects the quality of
the solution.  We derive the asymptotic variance
produced by SGD for a constant step size on a simple quadratic
model that obeys the variance bound given in (\ref{eqn:newmodel}).
We use objective $f(w) = \frac{1}{2} c w^2$ and
noisy gradient samples of the form
$
  \nabla \tilde f(w)
  =
  c w - \tilde b w - \tilde h,
$
where $\tilde b$ and $\tilde h$ are independent random variables with mean $0$
and variance $\beta^2$ and $\sigma^2$, respectively.  At each step, our algorithm
will run the standard SGD update rule, with constant step size $\alpha$,
independently in each of $M$ worker threads.  Then, it will choose to average
all the models with probability $\zeta$, i.e., expected phase length is 
$\zeta^{-1}$. Under these conditions, we can explicitly derive the asymptotic
variance of the average of the models. We defer the proof of
this result to Appendix~\ref{sec:lemmaStochasticAvgProof}.
\begin{lemma}
\label{lemmaStochasticAvg}
The asymptotic variance of the average in the above algorithm is
\[
	\lim_{t \rightarrow \infty}
	\mathbf{Var}\left( \frac{1}{M} \sum_{i=1}^M w_{i,t} \right)
	=
	\frac{\alpha \sigma^2}{M}
  \left(
    2 c - \alpha c^2 - \alpha \beta^2 \frac{1 + \eta M^{-1}}{1 + \eta}
  \right)^{-1}
\]
where $ \eta = \frac{
    \zeta
  }{
    (1 - \zeta)
    \alpha (2 c - \alpha c^2)
  }$.
\end{lemma}
This lemma shows that averaging can have asymptotic benefits; we will see in our experiments that those benefits can, in some cases, be observable in the non-asymptotic case too.

\subsection{Averaging for Non-Convex Problems}

The analysis of averaging becomes more complicated in the case of non-convex
problems.  Unlike in the convex case, where all worker threads are guaranteed
to converge independently to a single optimum, in non-convex optimization
the algorithm may have multiple stable fixed points.  Because different workers
may converge to different points,
one-shot averaging can produce incorrect results, even if we use an unbounded
number of gradient samples.  Averaging for non-convex optimization has been
previously studied in some settings~\cite{lian2015nonconvex,zhang2015splash},
but the effect of the frequency of averaging among workers has not been
considered.  We can illustrate this effect with a simple example.

Consider minimizing the function $f(w) = (w^2 - 1)^2$ by using gradient samples
of the form $\nabla \tilde f(w) = 4 (w^3 - w + \tilde u)$, where $\tilde u$
is a standard normal random variable.  This problem is a one-dimensional
version of the ubiquitous matrix completion problem, on which single-threaded
SGD is known to work~\cite{desa2015global}.  We ran SGD on this problem using
step
size $\alpha = 0.025$ and 10000 steps, parallelizing over 24 workers.
Our goal is to test the effect of various averaging rates on this non-convex
problem.  Here, one-shot averaging achieves an 
abysmal average objective of $0.922$.  Averaging even
$0.1\%$ of the time improves this to $0.274$, and averaging after $10\%$ of
the steps achieves an objective of $0.011$ --- a much more satisfying result.
These results suggest that properly selecting the averaging rate is even
more important in the non-convex setting.

\begin{wrapfigure}{r}{.35\linewidth}
\includegraphics[width=\linewidth]{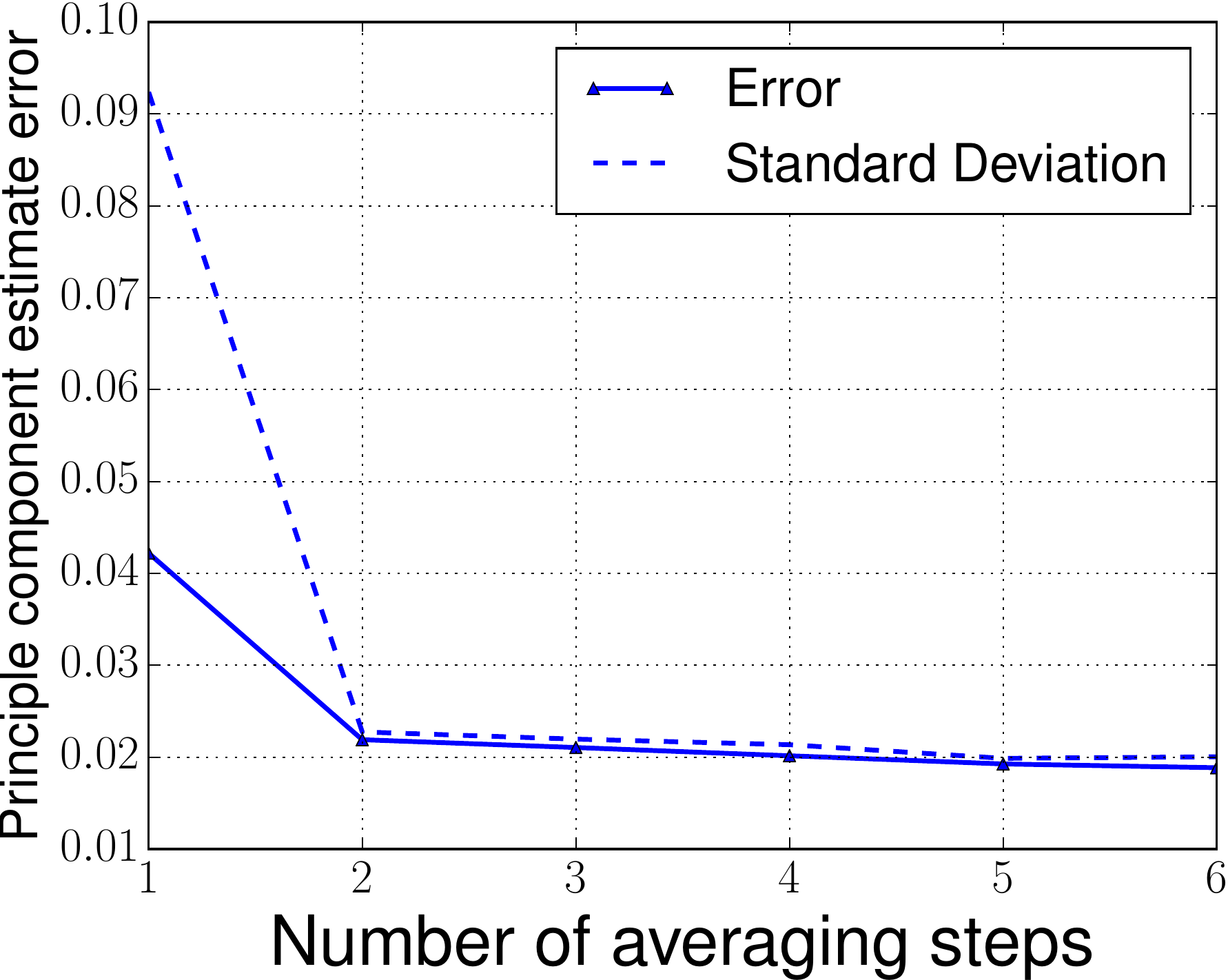}
\caption{PCA}
\label{fig:pca}
\end{wrapfigure} 

\textbf{PCA}
To demonstrate this point on an real problem, we run a simple simulation of Principal Component Analysis. We generate samples from a $20$ dimensional, zero-mean Gaussian with spectrum $[1.0, 0.7, \ldots 0.7]$ and principal component $v_1$. We use Oja's update $w_{ik} = w_{i(k-1)} + \alpha x_{ik} x_{ik}^\top w_{i(k-1)}$, where $x_{ik}$ is the sample used by worker $i$ at time $k$. We simulate $48$ workers and give each $10^4$ random samples.  Figure~\ref{fig:pca} reports the principal component error, $1-|w^Tv_1|/(\|w\|\|v_1\|)$, as a function of the total number of averaging steps used throughout. One-shot averaging corresponds to the leftmost point. This practical result illustrates that one-shot averaging is not suited for non-convex problems, an issue resolved by more frequent averaging.

\section{Experiments}
\label{sec:experiments}
In this section, we empirically investigate the effect of averaging in both convex problems and convolutional neural networks, on synthetic and canonical datasets. These results provide support for our intuition and theory discussed in the previous section.

\subsection{Convex Problems}
\textbf{Datasets and setup.}
We conduct experiments on three datasets for least squares regression and two for logistic regression. 
All the datasets in Table~\ref{tab:dataset} are available through the lib-svm dataset hub \cite{libsvmdataset}.
We use a four-socket NUMA machine
with threads balanced on the four sockets and bound to physical cores without hyperthreading. 

\begin{figure}[t]
	\centering
	\begin{subfigure}[b]{0.32\linewidth}
	\centering
		\includegraphics[width=\linewidth]{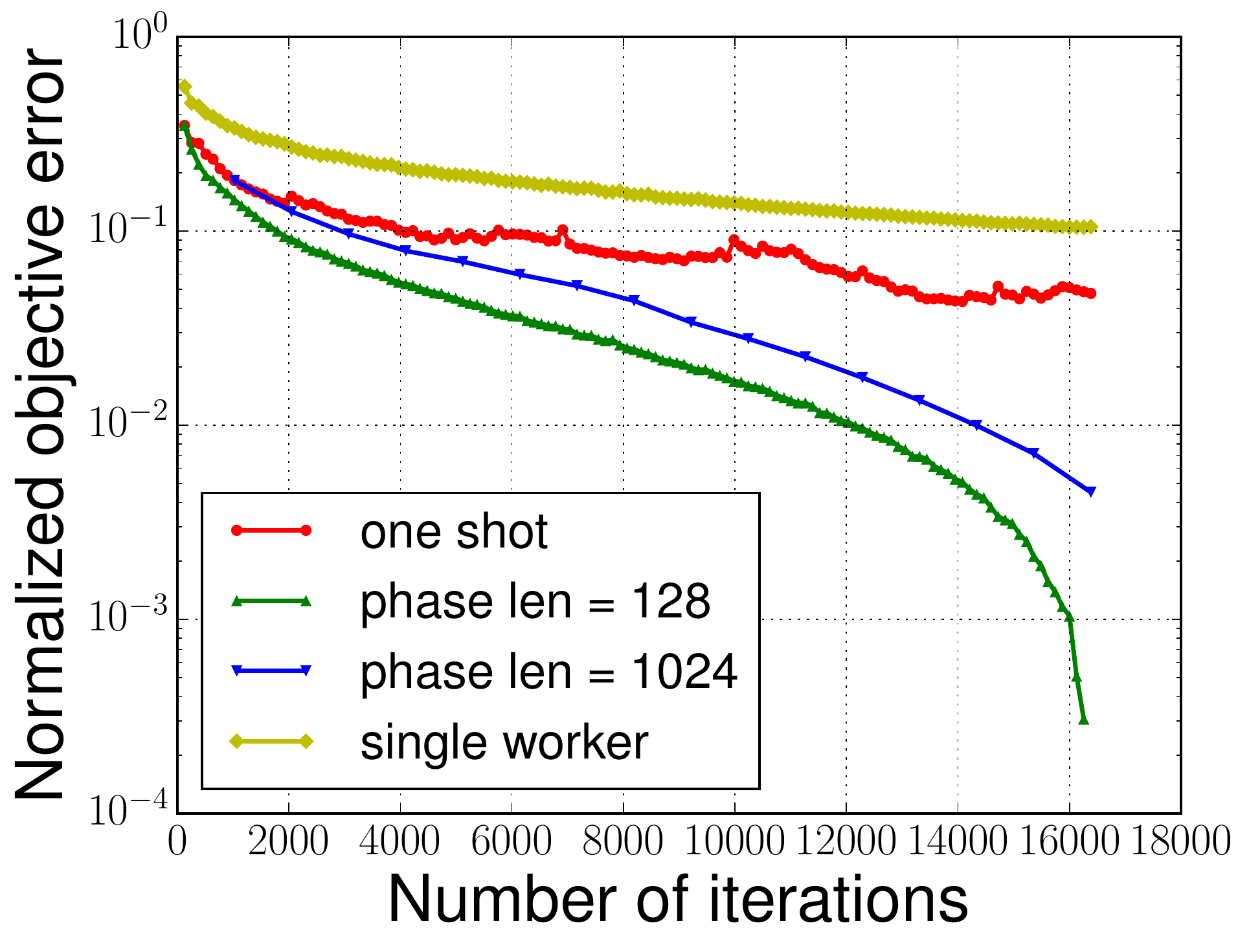} 
		\caption{E2006-TfIdf}
		\label{fig:E2006TfIdfGrid}	
	\end{subfigure}
	\begin{subfigure}[b]{0.32\linewidth}
	\centering
		\includegraphics[width=\linewidth]{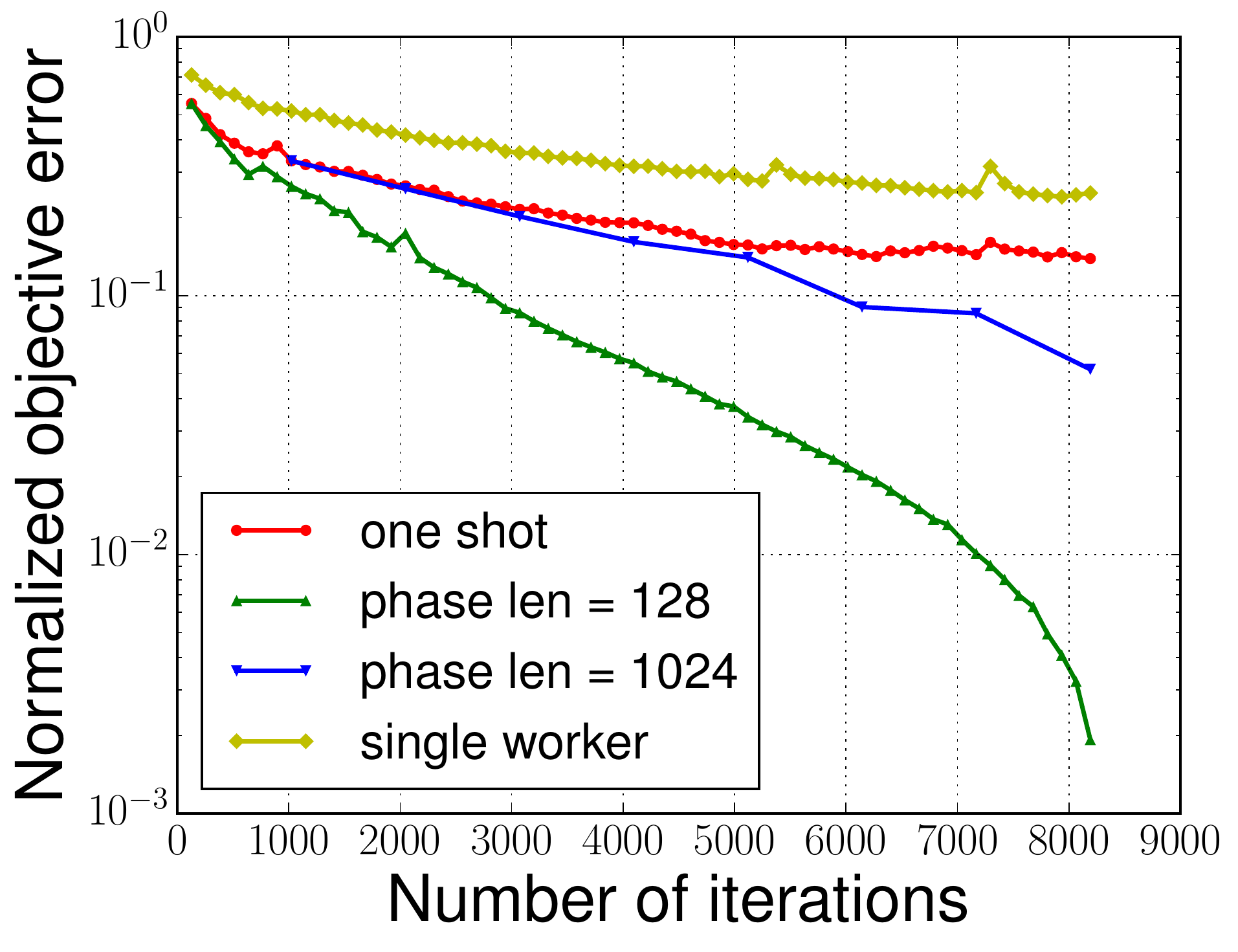} 
		\caption{E2006-log1p}
		\label{fig:E2006Log1pGrid}	
	\end{subfigure}
	\begin{subfigure}[b]{0.32\linewidth}
	\centering
		\includegraphics[width=\linewidth]{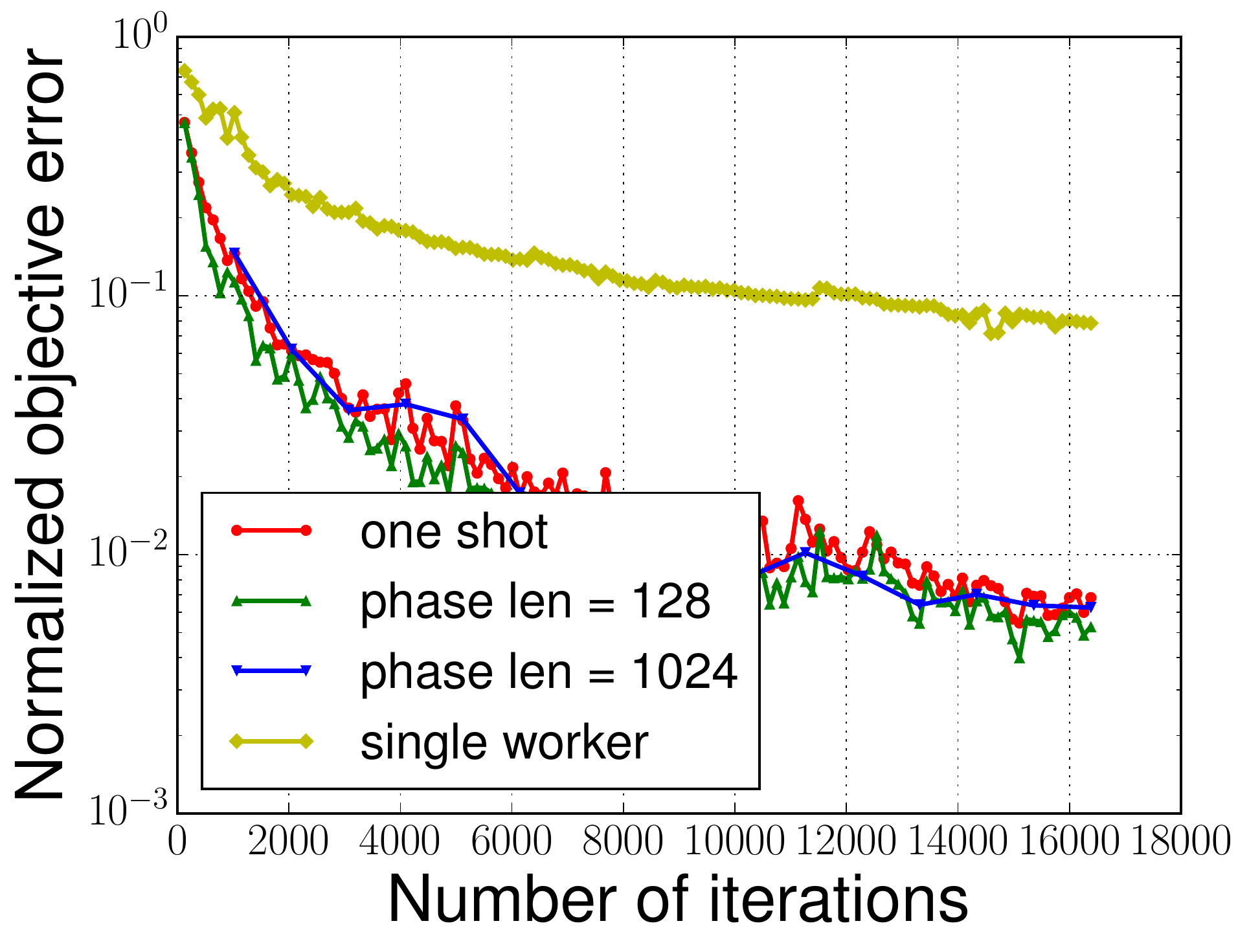} 
		\caption{YearPrediction}
		\label{fig:YPMSDGrid}	
	\end{subfigure}
	\begin{subfigure}[b]{0.32\linewidth}
	\centering
		\includegraphics[width=\linewidth]{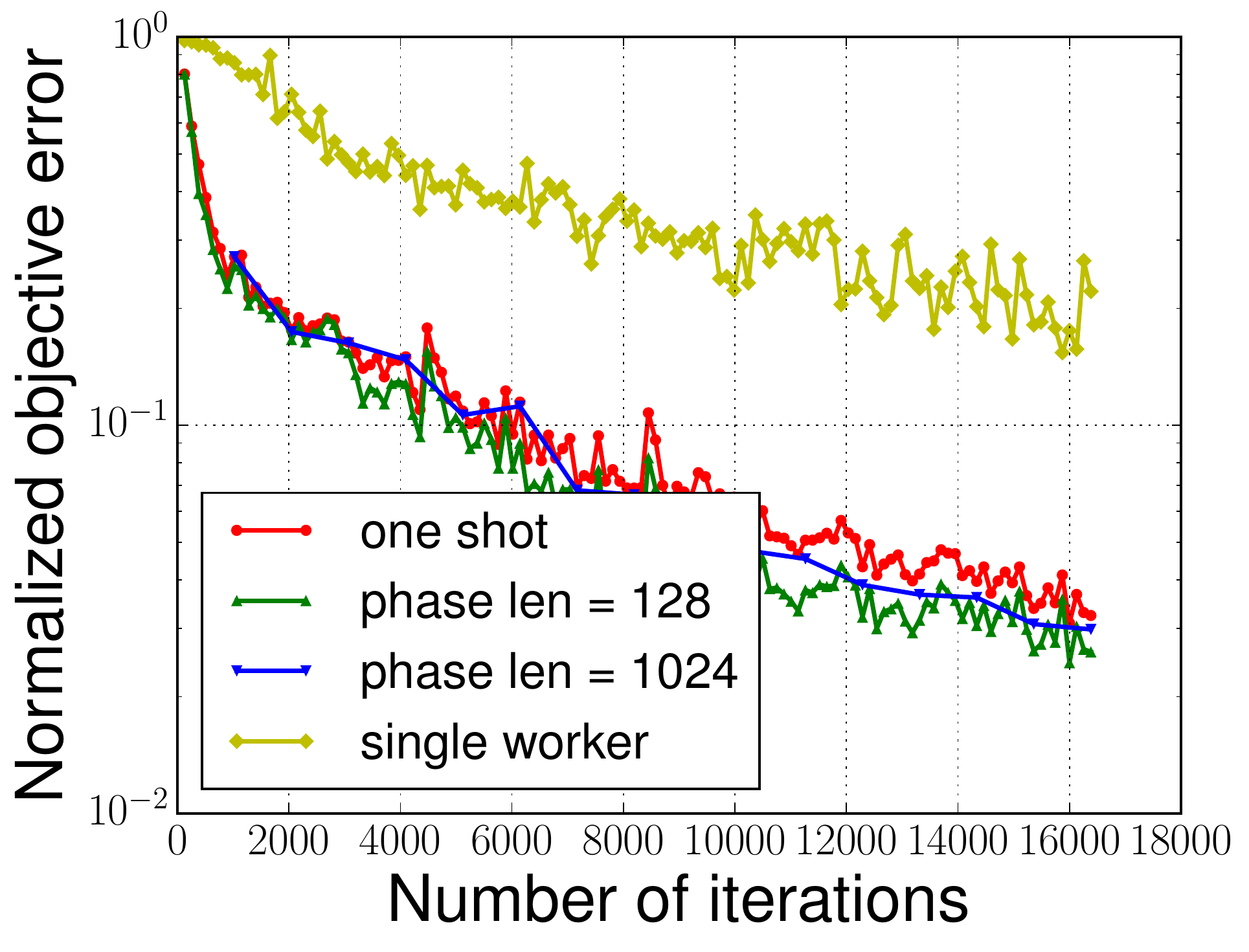} 
		\caption{HIGGS}
		\label{fig:HIGGSGrid}	
	\end{subfigure}
	\begin{subfigure}[b]{0.32\linewidth}
	\centering
		\includegraphics[width=\linewidth]{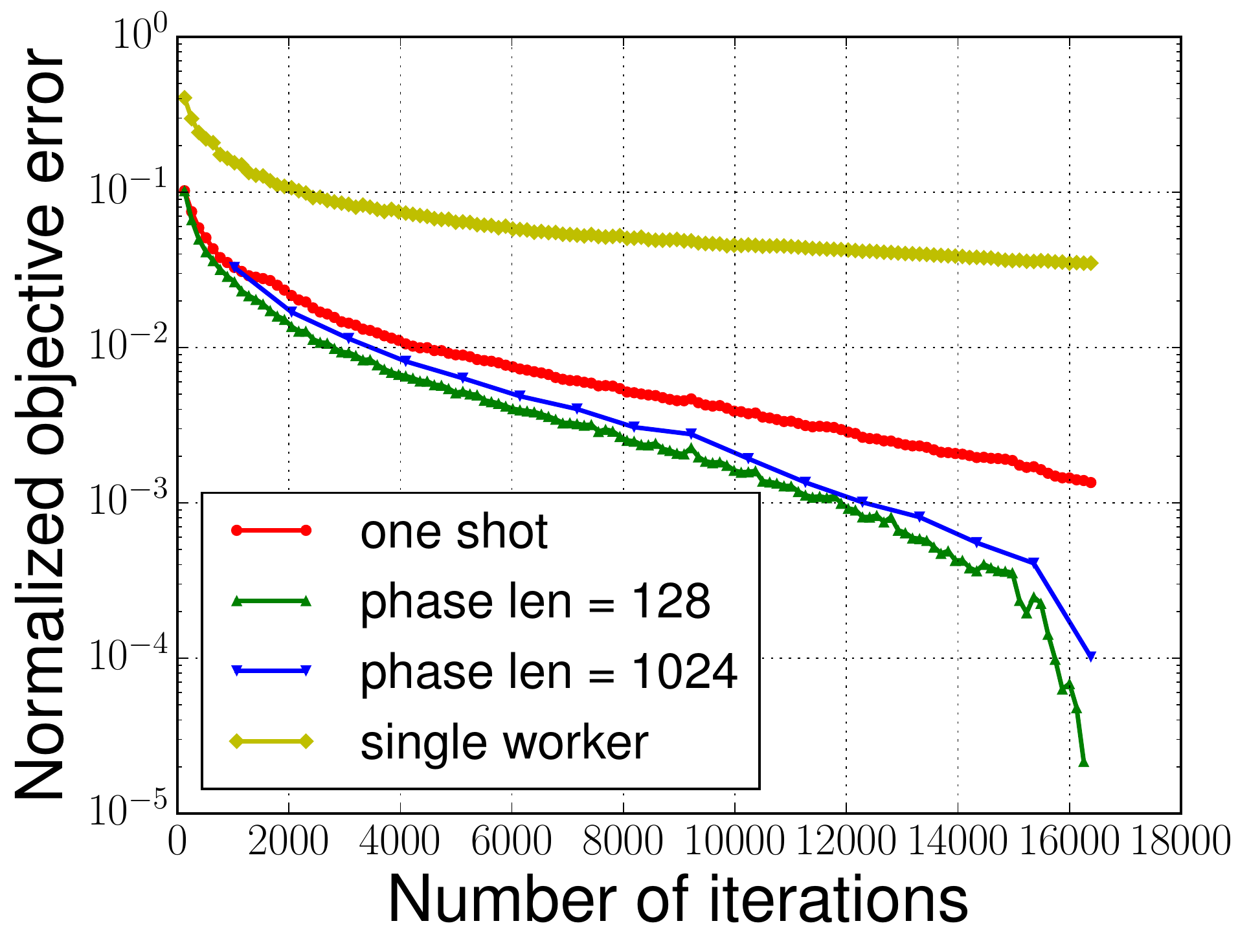} 
		\caption{RCV1}
		\label{fig:RCV1Grid}	
	\end{subfigure}
	\begin{subfigure}[b]{0.32\linewidth}
	\centering
		\begin{footnotesize}
		\begin{tabular}{c c}
		\toprule
		Data & Speedup \\
		\midrule
		E2006-tfidf & \textbf{2.26 x} \\
		E2006-log1p & \textbf{ >3 x} \\
		YearPrediction & \textbf{1.22 x} \\
		HIGGS & \textbf{1.32 x} \\
		RCV1 & \textbf{1x} \\
		\bottomrule
		\end{tabular}
		\end{footnotesize}
		\vspace{0.15in}
		\caption{Speedup to get 0.1 error.}
	\end{subfigure}
	\caption{Normalized suboptimality over time for convex experiments and principle component estimate error of the PCA example. The speedup column compares how fast the algorithm achieves normalized objective error 0.1. It compares averaging every 128 iterations to one shot averaging.}
	\label{fig:Results}
\end{figure}


\begin{table}[t]
\centering
\begin{scriptsize}
	\begin{tabular}{ c@{\hskip 1em}  c@{\hskip 1em}  c@{\hskip 1em}  c@{\hskip 1em}  c@{\hskip 1em} c}
	\toprule
	Dataset &  Model & \# Samples & \# Dimensions & $\sigma^2 $& $\rho \triangleq \beta^2 \|w_0 - w^\star\|^2 / \sigma^2$\\
	\midrule
	E2006-tfidf & LS &16,087 & 4,272,227 & $4.918 \times10^{-7}$ & $4.046 \times10^{9}$ \\ 
	E2006-log1p & LS & 16,087 & 150,360 &  $4.265 \times10^{-8}$ & $8.032 \times10^{8}$\\ 
	YearPrediction & LS & 463,715 & 90 & $94.151$& $2.865$\\ 
	HIGGS & LR & 11,000,000 & 28 & $7.324$ & $6.640$\\ 
	rcv1-binary & LR & 677,399 & 47,236 & $8.436 \times 10^{-9}$ & $1.558$ \\ 
	\bottomrule
	\end{tabular}
	\vspace{0.3em}
	\caption{Datasets, models (LS for least squares and LR for logistic regression) and related measurements. }
	\label{tab:dataset}
\end{scriptsize}
\end{table}

We run SGD with step size $\alpha / \left(t + d \right)$ where $t$ is the iteration index. Each experiment is repeated with three different data shuffles and we report the average objective at each iteration\footnote{We run only one experiment for E2206-log1p due to time limits.}.
For a fair comparison, we grid-search $\alpha$ and $d$, and at each iteration, we report the minimum average objective from the grid.
The objective is normalized so that the initial value at the initialization point is $1$ and the optimal value is $0$. 
We use $24$ workers to compare one-shot to periodic averaging (every $128$ steps and every $1024$ steps). We also report single worker results for context. 

\textbf{Measuring $\beta^2$ and $\sigma^2$.}
The variance model suggests that when $\rho$ is small, variance is dominated by the constant term, $\sigma^2$. In that case we know (cf.\ Section~\ref{sec:variance-model}) that periodic averaging yields little or no measurable benefits. On the other hand, we expect to see quicker convergence for periodic averaging when $\rho$ is large. To test this intuition, we measure all the quantities involved. For each experiment, (1) we find the approximate optimizer $w^*$; (2) we measure gradient variance at the optimum, which yields the value of $\sigma^2$; (3) we draw a random line that passes through $w^*$; (4) we take $9$ measurements of gradient variance along the line; (5) we calculate curvature based on those measurements; and (6) we repeat from Step 3 a number of times and average the measured curvatures, which yields an estimate for $\beta^2$.  We use these estimates to calculate the $\rho$ values reported in Table~\ref{tab:dataset}.


\textbf{Discussion.}
We see in Figures~\ref{fig:E2006TfIdfGrid}, \ref{fig:E2006Log1pGrid}, and \ref{fig:YPMSDGrid} significant gaps between periodic and one-shot averaging for sparse E2006-tfidf and E2006-log1p, while the gap is minor for the dense YearPrediction dataset. 
This is consistent with our model, as in Table~\ref{tab:dataset} we observe $\rho$ is considerably larger for sparse E2006-tfidf and E2006-log1p than dense YearPrediction. The gap becomes smaller for logistic regression on both sparse and dense data as shown in Figures~\ref{fig:HIGGSGrid} and \ref{fig:RCV1Grid}. Overall, we observe the anticipated correlation between speedup and measured value of $\rho$. 
The results demonstrate the finite-step effect of periodic averaging, thus complementing
our limiting analysis.


\subsection{Non-convex Problems}
\textbf{Architecture and setup.} We implement a two layer convolutional neural network using TensorFlow~\cite{tensorflow2015} on the digit recognition dataset MNIST. The dataset contains 60,000 samples for training and 10,000 for testing. We use a LeNet5-like architecture. More specifically, we use 32 and 64 $5\times5$ filters for the first and second convolution layers respectively. Each convolution layer is followed by a ReLu layer and a max-pooling layer with stride 2. Two fully connected layers are additionally inserted before the standard cross-entropy loss. We use momentum SGD with initial learning rate 0.01 and momentum coefficient 0.9. The step-size decays with a factor of 0.95 after each pass of training set. 

We deploy 4 workers with mini-batch size 8. Each worker uses a different data permutation. In addition to monitoring one-shot and periodic averaging with phase length 10, we also record the performance of single workers. For efficiency, we pick the workers with the best and worst loss on a subset of 5,000 training samples at the end of each phase. The performance of these two workers is then reported on the full training and test set.   

\begin{figure}[htb]
\centering
\begin{subfigure}[t]{.3\linewidth}
	\includegraphics[width=\linewidth]{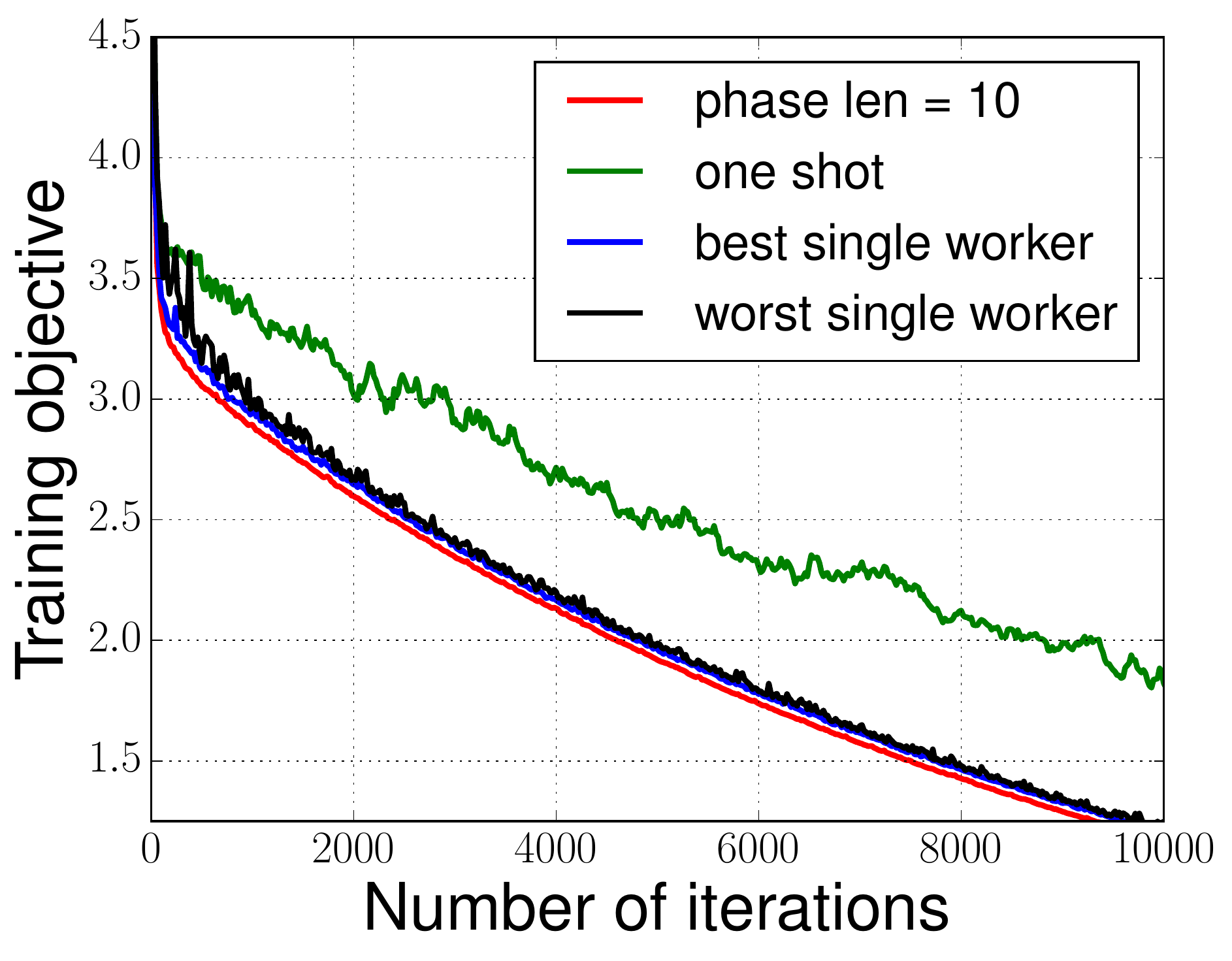}
	\caption{Training loss}
	\label{fig:TrainLoss}
\end{subfigure}
\begin{subfigure}[t]{.3\linewidth}
	\includegraphics[width=\linewidth]{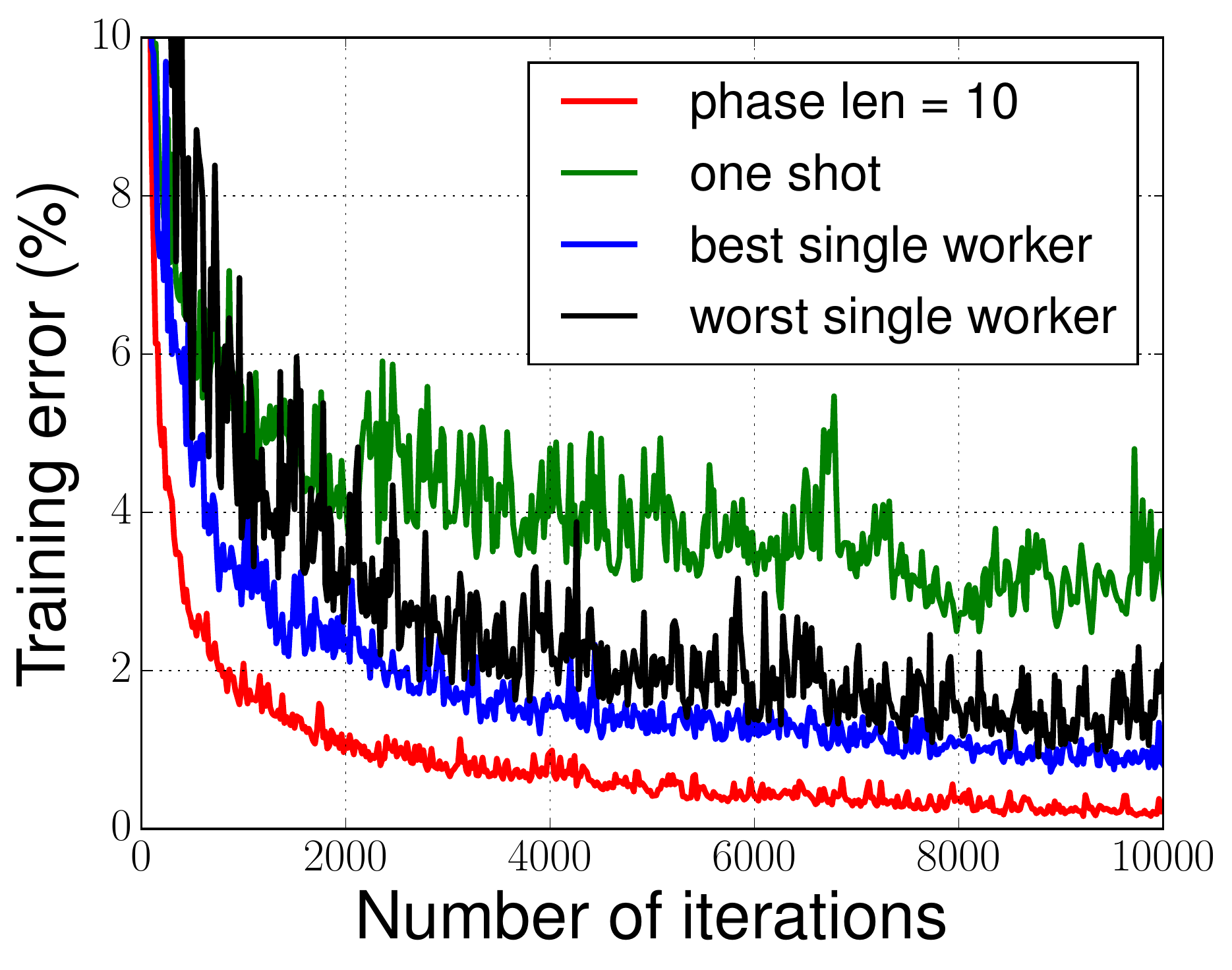}
	\caption{Training error}
	\label{fig:TrainErr}
\end{subfigure}
\begin{subfigure}[t]{.3\linewidth}
	\includegraphics[width=\linewidth]{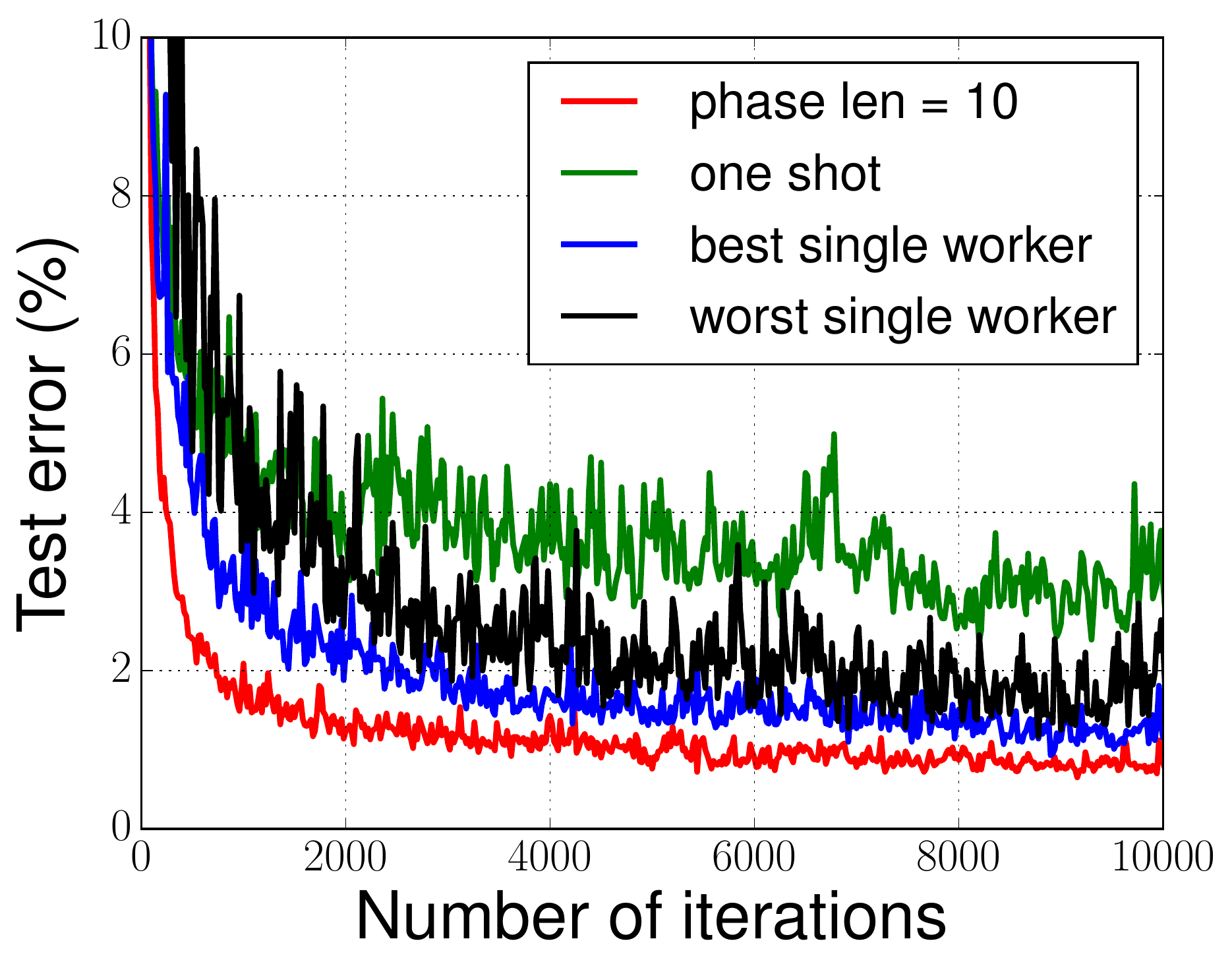}
	\caption{Test error}
	\label{fig:TestErr}
\end{subfigure}
\caption{Effect of averaging for CNN on MNIST. One-shot averaging degrades performance compared to single worker performance while periodic averaging improves the results.}
\label{fig:MNISTRes}
\end{figure}

\textbf{Discussion.} Figure~\ref{fig:MNISTRes} shows results on training loss, training error and test error with respect to the number of iterations. We see that periodic averaging can speed up convergence significantly compared to one-shot averaging, but also compared to the best single worker result. 
For the former, intuition is that one-shot averaging will take an average of different local minima. Unlike in convex settings, a single average in the end increases the training loss by improperly combining solutions. We observe that the relative performance of one-shot averaging worsens as the number of iterations increases. In particular, one-shot averaging is typically worse than the worst single worker's result. For the latter, we note that periodic averaging performs better than the best single worker, as it reduces variance.
%
%
%
%

\section{Conclusions}
\label{sec:conc}
In this paper, we investigate the effect of model averaging on parallel SGD in both convex and non-convex settings. Inspired by the analysis of quadratic problems, we propose a novel gradient variance model. The model gives intuition on the empirical benefits of periodic averaging on convex objectives, such as least squares regression and logistic regression. In the non-convex setting, we observe that
one-shot averaging can produce worse results, as it may combine 
estimates from different local minima.
In contrast, periodic averaging tends to bring the workers' estimates to a common basin of attraction. This leads to improved solution quality. Finally, due to its variance reduction properties, periodic averaging usually outperforms the best worker.

%


\bibliography{workshop}
\bibliographystyle{plainnat}

\appendix
%
%
%
%
%
\section{Proof of~\cref{lemmaStochasticAvg}}
\label{sec:lemmaStochasticAvgProof}

Assume that our task is minimizing the function
\[
  f(w) = \frac{1}{2} c w^2
\]
by using gradient samples of the form
\[
  \nabla \tilde f(w)
  =
  c w - \tilde b w - \tilde h,
\]
where $\tilde b$ and $\tilde h$ are independent $0\text{-mean}$ random variables with variance $\beta$ and $\gamma$ respectively. The
update rule is
\[
  w_{t+1} = (1 - \alpha c) w_t + \alpha (\tilde b w_t + \tilde h).
\]
Assume we initialize $w_0$ such that $\Exv{w_0} = 0$, which implies
$\Exv{w_t} = 0$, and we run the
update rule in $M$ independent threads. We denote the variance of the average
value of $w$ as
\[
  Q_t
  =
  \Exv{
    \left(
      \frac{1}{M} \sum_{i=1}^M
      w_{k, t+1}
    \right)^2
  }.
\]
We can write $Q_{t+1}$ as
\begin{equation}
\begin{aligned}
  Q_{t+1}
  &=
  \Exv{
    \left(
      \frac{1}{M} \sum_{i=1}^M
      w_{k, t+1}
    \right)^2
  } \\
  &=
  \Exv{
    \left(
      \frac{1}{M} \sum_{i=1}^M
      (1 - \alpha c) w_{i,t} + \alpha (\tilde b_{i,t} w_{i,t} + \tilde h_{i,t})
    \right)^2
  } \\
  &=
  \Exv{
    \left(
      \frac{1}{M} \sum_{i=1}^M
      (1 - \alpha c) w_{i,t})
    \right)^2
  }
  +
  \Exv{
    \left(
      \frac{1}{M} \sum_{i=1}^M
      \alpha \tilde b_{i,t} w_{i,t}
    \right)^2
  }
  +
  \Exv{
    \left(
      \frac{1}{M} \sum_{i=1}^M
      \alpha \tilde h_{i,t}
    \right)^2
  } \\
  &=
  (1 - \alpha c)^2
  \Exv{
    \left(
      \frac{1}{M} \sum_{i=1}^M
      w_{i,t})
    \right)^2
  }
  +
  \frac{\alpha^2}{M^2} \sum_{i=1}^M
  \Exv{\tilde b_{i,t}^2 w_{i,t}^2}
  +
  \frac{\alpha^2}{M^2} \sum_{i=1}^M
  \Exv{\tilde h_{i,t}^2} \\
  &=
  (1 - \alpha c)^2 Q_t
  +
  \frac{\alpha^2 \beta}{M^2} \sum_{i=1}^M
  \Exv{w_{i,t}^2}
  +
  \frac{\alpha^2 \gamma}{M}.
\end{aligned}
\label{equ:DecompQ}
\end{equation}
In addition, we also have that for any $i$,
\begin{dmath*}
  \Exv{w_{i,t+1}^2}
  =
  \Exv{\left(
    (1 - \alpha c) w_{i,t} + \alpha (\tilde b w_{i,t} + \tilde h)
  \right)^2}
  =
  (1 - \alpha c)^2 \Exv{w_{i,t}^2}
  +
  \alpha^2 \Exv{\tilde b^2 w_{i,t}^2}
  +
  \alpha^2 \Exv{\tilde h^2}
  =
  (1 - \alpha c)^2 \Exv{w_{i,t}^2}
  +
  \alpha^2 \beta \Exv{w_{i,t}^2}
  +
  \alpha^2 \gamma.
\end{dmath*}
As we further assume that $\Exv{w_{i,0}^2}$ is independent of $i$, let
\[
  P_t = \Exv{w_{i,t}^2},
\]
we can conclude that
\begin{equation}
  P_{t+1}
  =
  (1 - \alpha c)^2 P_t
  +
  \alpha^2 \beta P_t
  +
  \alpha^2 \gamma.
  \label{equ:PNoAve}
\end{equation}
By substituting $P_t$ into Equation~\eqref{equ:DecompQ}, we also have
\begin{equation}
  Q_{t+1}
  =
  (1 - \alpha c)^2 Q_t
  +
  \frac{\alpha^2 \beta}{M} P_t
  +
  \frac{\alpha^2 \gamma}{M}.
  \label{equ:QNoAve}
\end{equation}
Equation~\eqref{equ:PNoAve} and~\eqref{equ:QNoAve} determine the evolution of the system in the case when
averaging does not happen. In the case where averaging happens,
our update rule is
\[
  w_{i,t+1} = \frac{1}{M} \sum_{i=1}^M w_{i,t}
\]
which results in
\[
  Q_{t+1} = Q_t
\]
and
\[
  P_{t+1} = Q_t.
\]
If we choose to average at each timestep with probability $\zeta$, then by
the law of total expectation,
\[
  \left[\begin{array}{c}
    Q_{t+1} \\ P_{t+1}
  \end{array}\right]
  =
  (1 - \zeta)
  \left[\begin{array}{c}
    (1 - \alpha c)^2 Q_t + \frac{\alpha^2 \beta}{M} P_t + \frac{\alpha^2 \gamma}{M} \\
    (1 - \alpha c)^2 P_t + \alpha^2 \beta P_t + \alpha^2 \gamma
  \end{array}\right]
  +
  \zeta
  \left[\begin{array}{c}
    Q_t \\
    Q_t
  \end{array}\right].
\]
Based on the above equation, we want to find the asymptotic value of $Q_t$, i.e. the
value that produces a steady-state. To find the state, we need to
solve the linear equation
\[
  \left[\begin{array}{c}
    Q \\ P
  \end{array}\right]
  =
  (1 - \zeta)
  \left[\begin{array}{c}
    (1 - \alpha c)^2 Q + \frac{\alpha^2 \beta}{M} P + \frac{\alpha^2 \gamma}{M} \\
    (1 - \alpha c)^2 P + \alpha^2 \beta P + \alpha^2 \gamma
  \end{array}\right]
  +
  \zeta
  \left[\begin{array}{c}
    Q \\
    Q
  \end{array}\right].
\]
By subtracting from both sides and diving by $(1 - \zeta)$, we can write this
as
\[
  \left[\begin{array}{c}
    Q \\ P
  \end{array}\right]
  =
  \left[\begin{array}{c}
    (1 - \alpha c)^2 Q + \frac{\alpha^2 \beta}{M} P + \frac{\alpha^2 \gamma}{M} \\
    (1 - \alpha c)^2 P + \alpha^2 \beta P + \alpha^2 \gamma
  \end{array}\right]
  +
  \frac{\zeta}{1 - \zeta}
  \left[\begin{array}{c}
    0 \\
    Q - P
  \end{array}\right]
\]
which is equivalent to
\[
  (1 - (1 - \alpha c)^2)
  \left[\begin{array}{c}
    Q \\ P
  \end{array}\right]
  =
  \left[\begin{array}{c}
    \frac{\alpha^2 \beta}{M} P + \frac{\alpha^2 \gamma}{M} \\
    \alpha^2 \beta P + \alpha^2 \gamma
  \end{array}\right]
  +
  \frac{\zeta}{1 - \zeta}
  \left[\begin{array}{c}
    0 \\
    Q - P
  \end{array}\right].
\]
Let $\rho = (1 - (1 - \alpha c)^2)$ and
$\eta \rho = \frac{\zeta}{1 - \zeta}$,
we have
\[
  \rho
  \left[\begin{array}{c}
    M Q \\ P
  \end{array}\right]
  =
  \left[\begin{array}{c}
    \alpha^2 \beta P + \alpha^2 \gamma \\
    \alpha^2 \beta P + \alpha^2 \gamma
  \end{array}\right]
  +
  \eta \rho
  \left[\begin{array}{c}
    0 \\
    Q - P
  \end{array}\right],
\]
It can be presented more compactly in matrix form as
\[
  \left[\begin{array}{c c}
    \rho M & - \alpha^2 \beta \\
    -\eta \rho & \rho - \alpha^2 \beta + \eta \rho
  \end{array}\right]
  \left[\begin{array}{c}
    Q \\ P
  \end{array}\right]
  =
  \left[\begin{array}{c}
    \alpha^2 \gamma \\
    \alpha^2 \gamma
  \end{array}\right].
\]
The determinant of the left-most matrix is
\begin{align*}
  \left|\begin{array}{c c}
    \rho M & - \alpha^2 \beta \\
    -\eta \rho & \rho - \alpha^2 \beta + \eta \rho
  \end{array}\right|
  &=
  \rho M (\rho - \alpha^2 \beta + \eta \rho)
  -
  \alpha^2 \beta \eta \rho \\
  &=
  \rho^2 M - \alpha^2 \beta \rho M + \eta \rho^2 M
  -
  \alpha^2 \beta \eta \rho \\
  &=
  \rho^2 M (1 + \eta) - \alpha^2 \beta \rho (M + \eta).
\end{align*}
Thus we have
\begin{align*}
  Q
  &=
  \left[\begin{array}{c c}
    1 & 0
  \end{array}\right]
  \left[\begin{array}{c c}
    \rho M & - \alpha^2 \beta \\
    -\eta \rho & \rho - \alpha^2 \beta + \eta \rho
  \end{array}\right]^{-1}
  \left[\begin{array}{c}
    \alpha^2 \gamma \\
    \alpha^2 \gamma
  \end{array}\right] \\
  &= 
  \left[\begin{array}{c c}
    1 & 0
  \end{array}\right]
  \frac{
    1
  }{
    \rho^2 M (1 + \eta) - \alpha^2 \beta \rho (M + \eta)
  }
  \left[\begin{array}{c c}
    \rho - \alpha^2 \beta + \eta \rho & \alpha^2 \beta \\
    \eta \rho & \rho M
  \end{array}\right]
  \left[\begin{array}{c}
    \alpha^2 \gamma \\
    \alpha^2 \gamma
  \end{array}\right] \\
  &= 
  \alpha^2 \gamma
  \frac{
    \rho + \eta \rho
  }{
    \rho^2 M (1 + \eta) - \alpha^2 \beta \rho (M + \eta)
  } \\
  &= 
  \alpha^2 \gamma
  \left(
    \rho M - \alpha^2 \beta \frac{M + \eta}{1 + \eta}
  \right)^{-1} \\
  &=
  \alpha^2 \gamma
  \left(
    (2 \alpha c - \alpha^2 c^2) M - \alpha^2 \beta \frac{M + \eta}{1 + \eta}
  \right)^{-1} \\
  &=
  \frac{\alpha \gamma}{M}
  \left(
    2 c - \alpha c^2 - \alpha \beta \frac{1 + \eta M^{-1}}{1 + \eta}
  \right)^{-1}.
\end{align*}
This proves the result in the paper, for the chosen assignment of
\[
  \eta
  =
  \frac{
    \zeta
  }{
    (1 - \zeta)
    \alpha (2 c - \alpha c^2)
  }.
\]

\pagebreak

\end{document}